\documentclass{article}

\usepackage{graphicx} 
\usepackage{subcaption} 

\usepackage{natbib}

\usepackage{algorithm}
\usepackage{algorithmic}
\usepackage{hyperref}
\usepackage{booktabs}
\usepackage{multirow}
\usepackage{amsmath}

\newif\ifshowcomments
\showcommentstrue
\ifshowcomments
\newcommand\jirvin[1]{\textcolor{red}{[jirvin: #1]}}
\newcommand\pranav[1]{\textcolor{blue}{[pranav: #1]}}
\else
\newcommand\jirvin[1]{}
\newcommand\pranav[1]{}
\fi
\usepackage{array}

\usepackage[accepted]{icml2017}

\icmltitlerunning{CheXNet: Radiologist-Level Pneumonia Detection on Chest X-Rays with Deep Learning}
\title{Chest}

\begin{document} 

\twocolumn[
\icmltitle{CheXNet: Radiologist-Level Pneumonia Detection on Chest X-Rays with Deep Learning}

\icmlsetsymbol{equal}{*}

\begin{icmlauthorlist}
\icmlauthor{Pranav Rajpurkar}{equal,cs}
\icmlauthor{Jeremy Irvin}{equal,cs}
\icmlauthor{Kaylie Zhu}{cs}
\icmlauthor{Brandon Yang}{cs}
\icmlauthor{Hershel Mehta}{cs}
\icmlauthor{Tony Duan}{cs}
\icmlauthor{Daisy Ding}{cs}
\icmlauthor{Aarti Bagul}{cs}
\icmlauthor{Robyn L. Ball}{med}
\icmlauthor{Curtis Langlotz}{rad}
\icmlauthor{Katie Shpanskaya}{rad}
\icmlauthor{Matthew P. Lungren}{rad}
\icmlauthor{Andrew Y. Ng}{cs}
\end{icmlauthorlist}

\icmlaffiliation{cs}{Stanford University Department of Computer Science}
\icmlaffiliation{med}{Stanford University Department of Medicine}
\icmlaffiliation{rad}{Stanford University Department of Radiology}

\icmlcorrespondingauthor{Pranav Rajpurkar}{pranavsr@cs.stanford.edu}
\icmlcorrespondingauthor{Jeremy Irvin}{jirvin16@cs.stanford.edu}

\icmlkeywords{healthcare, medical imaging, convolutional neural networks, xrays, pneumonia}
\vskip 0.1in

]

\printAffiliationsAndNotice{\icmlEqualContribution}

\begin{abstract} 
We develop an algorithm that can detect pneumonia from chest X-rays at a level exceeding practicing radiologists. Our algorithm, CheXNet, is a 121-layer convolutional neural network trained on ChestX-ray14, currently the largest publicly available chest X-ray dataset, containing over 100,000 frontal-view X-ray images with 14 diseases. Four practicing academic radiologists annotate a test set, on which we compare the performance of CheXNet to that of radiologists. We find that CheXNet exceeds average radiologist performance on the F1 metric. We extend CheXNet to detect all 14 diseases in ChestX-ray14 and achieve state of the art results on all 14 diseases.
\end{abstract}

\begin{figure}[ht!]
  \centering
  \includegraphics[height=11cm]{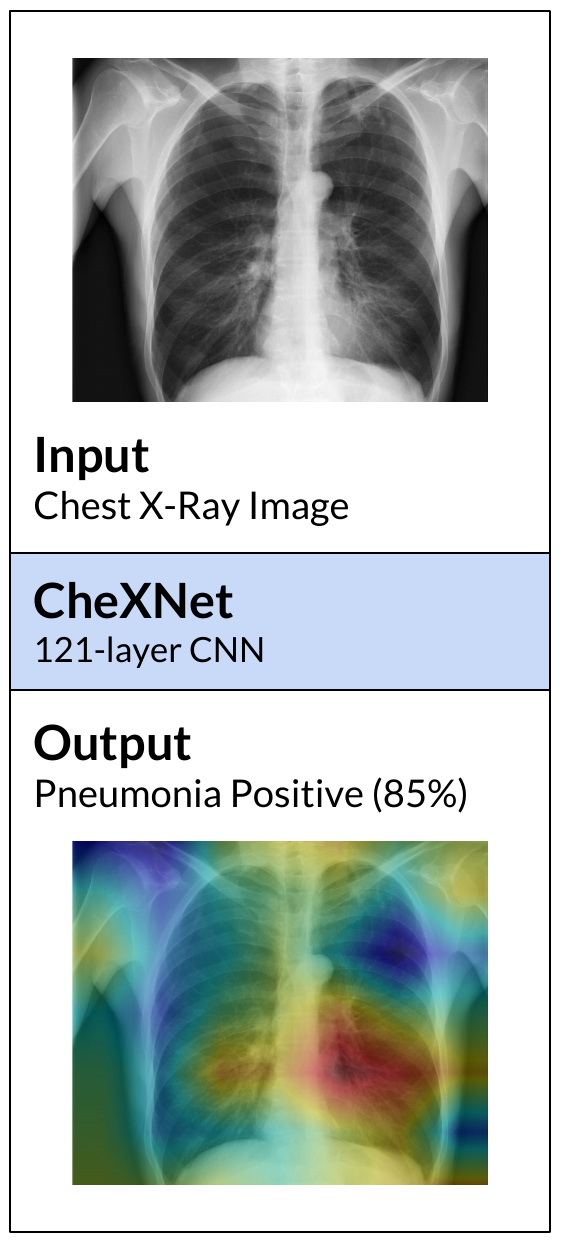}
  \caption{
       CheXNet is a 121-layer convolutional neural network that takes a chest X-ray image as input, and outputs the probability of a pathology. On this example, CheXnet correctly detects pneumonia and also localizes areas in the image most indicative of the pathology.
  }
  \label{fig:system}
\end{figure}

{\let\thefootnote\relax\footnote{Project website at \url{https://stanfordmlgroup.github.io/projects/chexnet}}}

\section{Introduction}
More than 1 million adults are hospitalized with pneumonia and around 50,000 die from the disease every year in the US alone \citep{pneumonia_2017}. Chest X-rays are currently the best available method for diagnosing pneumonia \citep{world2001standardization}, playing a crucial role in clinical care \citep{franquet2001imaging} and epidemiological studies \citep{cherian2005standardized}. However, detecting pneumonia in chest X-rays is a challenging task that relies on the availability of expert radiologists. In this work, we present a model that can automatically detect pneumonia from chest X-rays at a level exceeding practicing radiologists.

Our model, ChexNet (shown in Figure \ref{fig:system}), is a 121-layer convolutional neural network that inputs a chest X-ray image and outputs the probability of pneumonia along with a heatmap localizing the areas of the image most indicative of pneumonia. We train CheXNet on the recently released ChestX-ray14 dataset \citep{Wang2017}, which contains 112,120 frontal-view chest X-ray images individually labeled with up to 14 different thoracic diseases, including pneumonia. We use dense connections \citep{Huang2016} and batch normalization \citep{Ioffe2015} to make the optimization of such a deep network tractable.

Detecting pneumonia in chest radiography can be difficult for radiologists. The appearance of pneumonia in X-ray images is often vague, can overlap with other diagnoses, and can mimic many other benign abnormalities. These discrepancies cause considerable variability among radiologists in the diagnosis of pneumonia \citep{neuman2012variability, davies1996reliability, hopstaken2004inter}. To estimate radiologist performance, we collect annotations from four practicing academic radiologists on a subset of 420 images from ChestX-ray14. On these 420 images, we measure performance of individual radiologists and the model.

We find that the model exceeds the average radiologist performance on the pneumonia detection task. To compare CheXNet against previous work using ChestX-ray14, we make simple modifications to CheXNet to detect all 14 diseases in ChestX-ray14, and find that we outperform best published results on all 14 diseases. Automated detection of diseases from chest X-rays at the level of expert radiologists would not only have tremendous benefit in clinical settings, it would also be invaluable in delivery of health care to populations with inadequate access to diagnostic imaging specialists.

\begin{table}[t]
  
  \begin{center}
    \begin{tabular}{p{40mm} r}
      \toprule
      & F1 Score (95\% CI)\\
      \midrule
      Radiologist 1 & 0.383 (0.309, 0.453)\\
      Radiologist 2 & 0.356 (0.282, 0.428)\\
      Radiologist 3 & 0.365 (0.291, 0.435)\\
      Radiologist 4 & 0.442 (0.390, 0.492)\\
      \midrule
      Radiologist Avg. & 0.387 (0.330, 0.442)\\
      CheXNet & 0.435 (0.387, 0.481)\\
      \bottomrule
    \end{tabular}
  \end{center}
  \caption{We compare radiologists and our model on the F1 metric, which is the harmonic average of the precision and recall of the models. CheXNet achieves an F1 score of 0.435 (95\% CI 0.387, 0.481), higher than the radiologist average of 0.387 (95\% CI 0.330, 0.442). We use the bootstrap to find that the difference in performance is statistically significant.}
  \label{table:test_results}
\end{table}

\section{CheXNet}

\subsection{Problem Formulation}
The pneumonia detection task is a binary classification problem, where the input is a frontal-view chest X-ray image $X$ and the output is a binary label $y\in\{0, 1\}$ indicating the absence or presence of pneumonia respectively. For a single example in the training set, we optimize the weighted binary cross entropy loss
\begin{eqnarray*}
L(X, y) &=& -w_+\cdot y \log p(Y=1|X)\\
&&- w_-\cdot (1 - y)\log p(Y=0|X),
\end{eqnarray*}
where $p(Y=i|X)$ is the probability that the network assigns to the label $i$, $w_+=|N|/(|P|+|N|)$, and $w_-=|P|/(|P|+|N|)$ with $|P|$ and $|N|$ the number of positive cases and negative cases of pneumonia in the training set respectively.

\subsection{Model Architecture and Training}

CheXNet is a 121-layer Dense Convolutional
Network (DenseNet) \citep{Huang2016} trained on the ChestX-ray 14 dataset. DenseNets improve flow of information and gradients through the network, making the optimization of very deep networks tractable. We replace the final fully connected layer with one that has a single output, after which we apply a sigmoid nonlinearity. 

The weights of the network are initialized with weights from a model pretrained on ImageNet \cite{Deng2009}. The network is trained end-to-end using Adam with standard parameters ($\beta_1=0.9$ and $\beta_2=0.999$) \cite{Kingma2014}. We train the model using minibatches of size 16.
We use an initial learning rate of $0.001$ that is decayed by a factor of $10$ each time the validation loss plateaus after an epoch, and pick the model with the lowest validation loss.


\begin{table*}[ht!]    
\centering
\begin{tabular}{l c c c c}
\toprule
Pathology & Wang et al. (2017) & Yao et al. (2017) & CheXNet (ours) \\
\midrule
\midrule
 Atelectasis & 0.716 & 0.772    & \textbf{0.8094} \\
 Cardiomegaly & 0.807 & 0.904   & \textbf{0.9248} \\
 Effusion & 0.784 & 0.859       & \textbf{0.8638} \\
 Infiltration & 0.609 & 0.695   & \textbf{0.7345} \\
 Mass & 0.706 & 0.792           & \textbf{0.8676} \\
 Nodule & 0.671 & 0.717         & \textbf{0.7802} \\
 Pneumonia & 0.633 & 0.713     & \textbf{0.7680} \\
 Pneumothorax & 0.806 & 0.841  & \textbf{0.8887} \\
 Consolidation & 0.708 & 0.788 & \textbf{0.7901} \\
 Edema & 0.835 & 0.882         & \textbf{0.8878} \\
 Emphysema & 0.815 & 0.829     & \textbf{0.9371} \\
 Fibrosis & 0.769 & 0.767      & \textbf{0.8047} \\
 Pleural Thickening & 0.708 & 0.765  & \textbf{0.8062} \\
 Hernia & 0.767 & 0.914        & \textbf{0.9164} \\
\bottomrule
\end{tabular}
\caption{CheXNet outperforms the best published results on all 14 pathologies in the ChestX-ray14 dataset. In detecting Mass, Nodule, Pneumonia, and Emphysema, CheXNet has a margin of $>$0.05 AUROC over previous state of the art results.}
\label{tab:comparison}
\end{table*}

\section{Data}
\subsection{Training}
  We use the ChestX-ray14 dataset released by \citet{Wang2017} which contains 112,120 frontal-view X-ray images of 30,805 unique patients. \citet{Wang2017} annotate each image with up to 14 different thoracic pathology labels using automatic extraction methods on radiology reports. We label images that have pneumonia as one of the annotated pathologies as positive examples and label all other images as negative examples. For the pneumonia detection task, we randomly split the dataset into training (28744 patients, 98637 images), validation (1672 patients, 6351 images), and test (389 patients, 
420 images). There is no patient overlap between the sets.

Before inputting the images into the network, we downscale the images to $224\times 224$ and normalize based on the mean and standard deviation of images in the ImageNet training set. We also augment the training data with random horizontal flipping.

\subsection{Test}
We collected a test set of 420 frontal chest X-rays. Annotations were obtained independently from four practicing radiologists at Stanford University, who were asked to label all 14 pathologies in \citet{Wang2017}. The radiologists had 4, 7, 25, and 28 years of experience, and one of the radiologists is a sub-specialty fellowship trained thoracic radiologist. Radiologists did not have access to any patient information or knowledge of disease prevalence in the data. Labels were entered into a standardized data entry program.

\begin{figure*}[ht!]
\centering
\begin{subfigure}[t]{0.3\textwidth}
\includegraphics[width=0.95\textwidth]{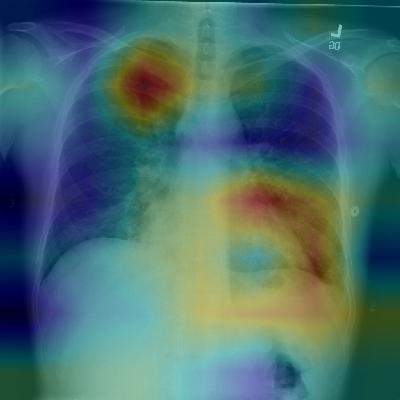}
\caption{Patient with multifocal community acquired pneumonia. The model correctly detects the airspace disease in the left lower and right upper lobes to arrive at the pneumonia diagnosis.}
\label{fig:cam_pneumonia}
\end{subfigure}%
\hfill
\begin{subfigure}[t]{0.3\textwidth}
\includegraphics[width=0.95\textwidth]{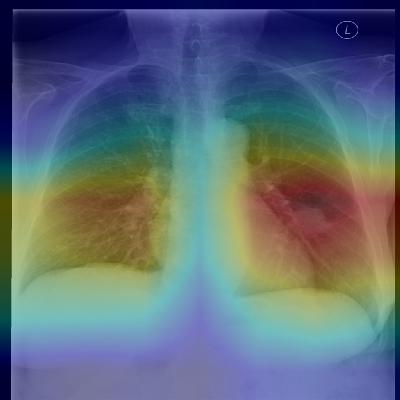}
\caption{Patient with a left lung nodule. The model identifies the left lower lobe lung nodule and correctly classifies the pathology.}
\label{fig:cam_nodule}
\end{subfigure}%
\hfill
\begin{subfigure}[t]{0.3\textwidth}
\includegraphics[width=0.95\textwidth]{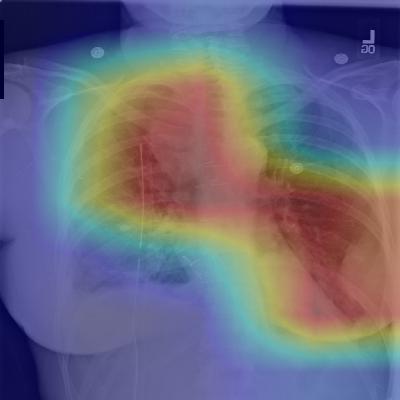}
\caption{Patient with primary lung malignancy and two large masses, one in the left lower lobe and one in the right upper lobe adjacent to the mediastinum. The model correctly identifies both masses in the X-ray.}
\label{fig:cam_mass}
\end{subfigure}%

\bigskip
\begin{subfigure}[t]{0.3\textwidth}
\includegraphics[width=0.95\textwidth]{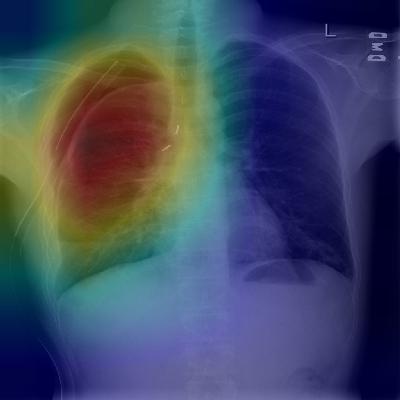}
\caption{Patient with a right-sided pneumothroax and chest tube. The model detects the abnormal lung to correctly predict the presence of pneumothorax (collapsed lung).}
\label{fig:cam_pneumothorax}
\end{subfigure}%
\hfill
\begin{subfigure}[t]{0.3\textwidth}
\includegraphics[width=0.95\textwidth]{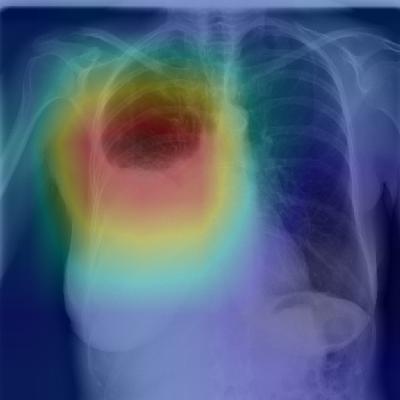} 
\caption{Patient with a large right pleural effusion (fluid in the pleural space). The model correctly labels the effusion and focuses on the right lower chest. }
\label{fig:cam_effusion}
\end{subfigure}%
\hfill
\begin{subfigure}[t]{0.3\textwidth}
\centering
\includegraphics[width=0.95\textwidth]{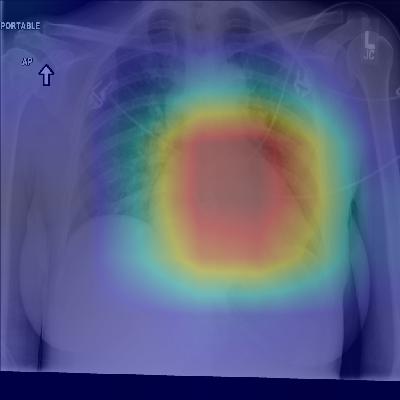}
\caption{Patient with congestive heart failure and cardiomegaly (enlarged heart). The model correctly identifies the enlarged cardiac silhouette. }
\label{fig:cam_cardiomegaly}
\end{subfigure}

\caption{CheXNet localizes pathologies it identifies using Class Activation Maps, which highlight the areas of the X-ray that are most important for making a particular pathology classification. The captions for each image are provided by one of the practicing radiologists.}
\label{fig:cams}
\end{figure*}

\section{CheXNet vs. Radiologist Performance}
\subsection{Comparison}
We assess the performance of both radiologists and CheXNet on the test set for the pneumonia detection task. Recall that for each of the images in the test set, we have 4 labels from four practicing radiologists and 1 label from CheXNet. We compute the F1 score for each individual radiologist and for CheXNet against each of the other 4 labels as ground truth. We report the mean of the 4 resulting F1 scores for each radiologist and for CheXNet, along with the average F1 across the radiologists. We use the bootstrap to construct 95\% bootstrap confidence intervals (CIs), calculating the average F1 score for both the radiologists and CheXNet on 10,000 bootstrap samples, sampled with replacement from the test set. We take the 2.5th and 97.5th percentiles of the F1 scores as the 95\% bootstrap CI. We find that CheXNet achieves an F1 score of 0.435 (95\% CI 0.387, 0.481), higher than the radiologist average of 0.387 (95\% CI 0.330, 0.442). Table \ref{table:test_results} summarizes the performance of each radiologist and of CheXNet.

To determine whether CheXNet's performance is statistically significantly higher than radiologist performance, we also calculate the difference between the average F1 score of CheXNet and the average F1 score of the radiologists on the same bootstrap samples. If the 95\% CI on the difference does not include zero, we conclude there was a significant difference between the F1 score of CheXNet and the F1 score of the radiologists. We find that the difference in F1 scores --- 0.051 (95\% CI 0.005, 0.084) --- does not contain 0, and therefore conclude that the performance of CheXNet is statistically significantly higher than radiologist performance.

\subsection{Limitations}
We identify three limitations of this comparison. 
First, only frontal radiographs were presented to the radiologists and model during diagnosis, but it has been shown that up to 15\% of accurate diagnoses require the lateral view \cite{raoof2012interpretation}; we thus expect that this setup provides a conservative estimate of performance. Third, neither the model nor the radiologists were not permitted to use patient history, which has been shown to decrease radiologist diagnostic performance in interpreting chest radiographs \cite{Berbaum1985,Potchen1979}; for example, given a pulmonary abnormality with a history of fever and cough, pneumonia would be appropriate rather than less specific terms such as infiltration or consolidation) \cite{Potchen1979}. 

\section{ChexNet vs. Previous State of the Art on the ChestX-ray14 Dataset}

We extend the algorithm to classify multiple thoracic pathologies by making three changes.
First, instead of outputting one binary label, ChexNet outputs a vector $t$ of binary labels indicating the absence or presence of each of the following 14 pathology classes: Atelectasis, Cardiomegaly, Consolidation, Edema, Effusion, Emphysema, Fibrosis, Hernia, Infiltration, Mass, Nodule, Pleural Thickening, Pneumonia, and Pneumothorax. 
Second, we replace the final fully connected layer in CheXNet with a fully connected layer producing a 14-dimensional output, after which we apply an elementwise sigmoid nonlinearity. The final output is the predicted probability of the presence of each pathology class.  Third, we modify the loss function to optimize the sum of unweighted binary cross entropy losses

\begin{eqnarray*}
L(X,y) &=& \sum_{c=1}^{14}[-y_c \log p(Y_c=1|X)\\
&&\hspace{1.7em}-(1 - y_c)\log p(Y_c=0|X)],
\end{eqnarray*}

where $p(Y_c=1|X)$ is the predicted probability that the image contains the pathology $c$ and $p(Y_c=0|X)$ is the predicted probability that the image does not contain the pathology $c$.

We randomly split the dataset into training (70\%), validation (10\%), and test (20\%) sets, following previous work on ChestX-ray14 \citep{Wang2017,Yao2017}. We ensure that there is no patient overlap between the splits. We compare the per-class AUROC of the model against the previous state of the art held by \citet{Yao2017} on 13 classes and \citet{Wang2017} on the remaining 1 class.

We find that CheXNet achieves state of the art results on all 14 pathology classes. Table~\ref{tab:comparison} illustrates the per-class AUROC comparison on the test set. On Mass, Nodule, Pneumonia, and Emphysema, we outperform previous state of the art considerably ($>0.05$ increase in AUROC).


\section{Model Interpretation}

To interpret the network predictions, we also produce heatmaps to visualize the areas of the image most indicative of the disease using class activation mappings (CAMs) \cite{Zhou2016}. To generate the CAMs, we feed an image into the fully trained network and extract the feature maps that are output by the final convolutional layer. Let $f_k$ be the $k$th feature map and let $w_{c,k}$ be the weight in the final classification layer for feature map $k$ leading to pathology $c$. We obtain a map $M_c$ of the most salient features used in classifying the image as having pathology $c$ by taking the weighted sum of the feature maps using their associated weights. Formally,
$$M_c=\sum_kw_{c,k}f_k.$$



We identify the most important features used by the model in its prediction of the pathology $c$ by upscaling the map $M_c$ to the dimensions of the image and overlaying the image.

Figure~\ref{fig:cams} shows several examples of CAMs on the pneumonia detection task as well as the 14-class pathology classification task. 

\section{Related Work}
Recent advancements in deep learning and large datasets have enabled algorithms to surpass the performance of medical professionals in a wide variety of medical imaging tasks, including diabetic retinopathy detection \cite{Gulshan2016}, skin cancer classification \cite{Esteva2017}, arrhythmia detection \citep{Rajpurkar2017}, and hemorrhage identification \cite{Grewal2017}.

Automated diagnosis from chest radiographs has received increasing attention with algorithms  for pulmonary tuberculosis classification \cite{lakhani2017deep2} and lung nodule detection \cite{Huang2017}. \citet{Islam2017} studied the performance of various convolutional architectures on different abnormalities using the publicly available OpenI dataset \citep{demner2015preparing}. \citet{Wang2017} released ChestX-ray-14, an order of magnitude larger than previous datasets of its kind, and also benchmarked different convolutional neural network architectures pre-trained on ImageNet. Recently \citet{Yao2017} exploited statistical dependencies between labels in order make more accurate predictions, outperforming \citet{Wang2017} on 13 of 14 classes.

\section{Conclusion}
Pneumonia accounts for a significant proportion of patient morbidity and mortality \cite{gonccalves2013community}. Early diagnosis and treatment of pneumonia is critical to preventing complications including death \cite{aydogdu2010mortality}. With approximately 2 billion procedures per year, chest X-rays are the most common imaging examination tool used in practice, critical for screening, diagnosis, and management of a variety of diseases including pneumonia \cite{raoof2012interpretation}. However, two thirds of the global population lacks access to radiology diagnostics, according to an estimate by the World Health Organization \cite{mollura2010white}. There is a shortage of experts who can interpret X-rays, even when imaging equipment is available, leading to increased mortality from treatable diseases \cite{kesselman20162015}.

We develop an algorithm which detects pneumonia from frontal-view chest X-ray images at a level exceeding practicing radiologists. We also show that a simple extension of our algorithm to detect multiple diseases outperforms previous state of the art on ChestX-ray14, the largest publicly available chest X-ray dataset. With automation at the level of experts, we hope that this technology can improve healthcare delivery and increase access to medical imaging expertise in parts of the world where access to skilled radiologists is limited.


\pagebreak
\section{Acknowledgements}
We would like to acknowledge the Stanford Center for Artificial Intelligence in Medicine and Imaging for clinical dataset infrastructure support (\url{AIMI.stanford.edu}). 

\bibliography{bibliography}

\begin{thebibliography}{29}
\providecommand{\natexlab}[1]{#1}
\providecommand{\url}[1]{\texttt{#1}}
\expandafter\ifx\csname urlstyle\endcsname\relax
  \providecommand{\doi}[1]{doi: #1}\else
  \providecommand{\doi}{doi: \begingroup \urlstyle{rm}\Url}\fi

\bibitem[Aydogdu et~al.(2010)Aydogdu, Ozyilmaz, Aksoy, Gursel, and
  Ekim]{aydogdu2010mortality}
Aydogdu, M, Ozyilmaz, E, Aksoy, Handan, Gursel, G, and Ekim, Numan.
\newblock Mortality prediction in community-acquired pneumonia requiring
  mechanical ventilation; values of pneumonia and intensive care unit severity
  scores.
\newblock \emph{Tuberk Toraks}, 58\penalty0 (1):\penalty0 25--34, 2010.

\bibitem[Berbaum et~al.(1985)Berbaum, Franken~Jr, and Smith]{Berbaum1985}
Berbaum, K, Franken~Jr, EA, and Smith, WL.
\newblock The effect of comparison films upon resident interpretation of
  pediatric chest radiographs.
\newblock \emph{Investigative radiology}, 20\penalty0 (2):\penalty0 124--128,
  1985.

\bibitem[CDC(2017)]{pneumonia_2017}
CDC, 2017.
\newblock URL \url{https://www.cdc.gov/features/pneumonia/index.html}.

\bibitem[Cherian et~al.(2005)Cherian, Mulholland, Carlin, Ostensen, Amin,
  Campo, Greenberg, Lagos, Lucero, Madhi, et~al.]{cherian2005standardized}
Cherian, Thomas, Mulholland, E~Kim, Carlin, John~B, Ostensen, Harald, Amin,
  Ruhul, Campo, Margaret~de, Greenberg, David, Lagos, Rosanna, Lucero, Marilla,
  Madhi, Shabir~A, et~al.
\newblock Standardized interpretation of paediatric chest radiographs for the
  diagnosis of pneumonia in epidemiological studies.
\newblock \emph{Bulletin of the World Health Organization}, 83\penalty0
  (5):\penalty0 353--359, 2005.

\bibitem[Davies et~al.(1996)Davies, Wang, Manson, Babyn, and
  Shuckett]{davies1996reliability}
Davies, H~Dele, Wang, Elaine E-l, Manson, David, Babyn, Paul, and Shuckett,
  Bruce.
\newblock Reliability of the chest radiograph in the diagnosis of lower
  respiratory infections in young children.
\newblock \emph{The Pediatric infectious disease journal}, 15\penalty0
  (7):\penalty0 600--604, 1996.

\bibitem[Demner-Fushman et~al.(2015)Demner-Fushman, Kohli, Rosenman, Shooshan,
  Rodriguez, Antani, Thoma, and McDonald]{demner2015preparing}
Demner-Fushman, Dina, Kohli, Marc~D, Rosenman, Marc~B, Shooshan, Sonya~E,
  Rodriguez, Laritza, Antani, Sameer, Thoma, George~R, and McDonald, Clement~J.
\newblock Preparing a collection of radiology examinations for distribution and
  retrieval.
\newblock \emph{Journal of the American Medical Informatics Association},
  23\penalty0 (2):\penalty0 304--310, 2015.

\bibitem[Deng et~al.(2009)Deng, Dong, Socher, Li, Li, and Fei-Fei]{Deng2009}
Deng, Jia, Dong, Wei, Socher, Richard, Li, Li-Jia, Li, Kai, and Fei-Fei, Li.
\newblock Imagenet: A large-scale hierarchical image database.
\newblock In \emph{Computer Vision and Pattern Recognition, 2009. CVPR 2009.
  IEEE Conference on}, pp.\  248--255. IEEE, 2009.

\bibitem[Esteva et~al.(2017)Esteva, Kuprel, Novoa, Ko, Swetter, Blau, and
  Thrun]{Esteva2017}
Esteva, Andre, Kuprel, Brett, Novoa, Roberto~A, Ko, Justin, Swetter, Susan~M,
  Blau, Helen~M, and Thrun, Sebastian.
\newblock Dermatologist-level classification of skin cancer with deep neural
  networks.
\newblock \emph{Nature}, 542\penalty0 (7639):\penalty0 115--118, 2017.

\bibitem[Franquet(2001)]{franquet2001imaging}
Franquet, T.
\newblock Imaging of pneumonia: trends and algorithms.
\newblock \emph{European Respiratory Journal}, 18\penalty0 (1):\penalty0
  196--208, 2001.

\bibitem[Gon{\c{c}}alves-Pereira et~al.(2013)Gon{\c{c}}alves-Pereira,
  Concei{\c{c}}{\~a}o, and P{\'o}voa]{gonccalves2013community}
Gon{\c{c}}alves-Pereira, Jo{\~a}o, Concei{\c{c}}{\~a}o, Catarina, and
  P{\'o}voa, Pedro.
\newblock Community-acquired pneumonia: identification and evaluation of
  nonresponders.
\newblock \emph{Therapeutic advances in infectious disease}, 1\penalty0
  (1):\penalty0 5--17, 2013.

\bibitem[Grewal et~al.(2017)Grewal, Srivastava, Kumar, and
  Varadarajan]{Grewal2017}
Grewal, Monika, Srivastava, Muktabh~Mayank, Kumar, Pulkit, and Varadarajan,
  Srikrishna.
\newblock Radnet: Radiologist level accuracy using deep learning for hemorrhage
  detection in ct scans.
\newblock \emph{arXiv preprint arXiv:1710.04934}, 2017.

\bibitem[Gulshan et~al.(2016)Gulshan, Peng, Coram, Stumpe, Wu, Narayanaswamy,
  Venugopalan, Widner, Madams, Cuadros, et~al.]{Gulshan2016}
Gulshan, Varun, Peng, Lily, Coram, Marc, Stumpe, Martin~C, Wu, Derek,
  Narayanaswamy, Arunachalam, Venugopalan, Subhashini, Widner, Kasumi, Madams,
  Tom, Cuadros, Jorge, et~al.
\newblock Development and validation of a deep learning algorithm for detection
  of diabetic retinopathy in retinal fundus photographs.
\newblock \emph{Jama}, 316\penalty0 (22):\penalty0 2402--2410, 2016.

\bibitem[Hopstaken et~al.(2004)Hopstaken, Witbraad, Van~Engelshoven, and
  Dinant]{hopstaken2004inter}
Hopstaken, RM, Witbraad, T, Van~Engelshoven, JMA, and Dinant, GJ.
\newblock Inter-observer variation in the interpretation of chest radiographs
  for pneumonia in community-acquired lower respiratory tract infections.
\newblock \emph{Clinical radiology}, 59\penalty0 (8):\penalty0 743--752, 2004.

\bibitem[Huang et~al.(2016)Huang, Liu, Weinberger, and van~der
  Maaten]{Huang2016}
Huang, Gao, Liu, Zhuang, Weinberger, Kilian~Q, and van~der Maaten, Laurens.
\newblock Densely connected convolutional networks.
\newblock \emph{arXiv preprint arXiv:1608.06993}, 2016.

\bibitem[Huang et~al.(2017)Huang, Park, Yan, Lee, Chu, Lin, Hussien, Rathmell,
  Thomas, Chen, et~al.]{Huang2017}
Huang, Peng, Park, Seyoun, Yan, Rongkai, Lee, Junghoon, Chu, Linda~C, Lin,
  Cheng~T, Hussien, Amira, Rathmell, Joshua, Thomas, Brett, Chen, Chen, et~al.
\newblock Added value of computer-aided ct image features for early lung cancer
  diagnosis with small pulmonary nodules: A matched case-control study.
\newblock \emph{Radiology}, pp.\  162725, 2017.

\bibitem[Ioffe \& Szegedy(2015)Ioffe and Szegedy]{Ioffe2015}
Ioffe, Sergey and Szegedy, Christian.
\newblock Batch normalization: Accelerating deep network training by reducing
  internal covariate shift.
\newblock In \emph{International Conference on Machine Learning}, pp.\
  448--456, 2015.

\bibitem[Islam et~al.(2017)Islam, Aowal, Minhaz, and Ashraf]{Islam2017}
Islam, Mohammad~Tariqul, Aowal, Md~Abdul, Minhaz, Ahmed~Tahseen, and Ashraf,
  Khalid.
\newblock Abnormality detection and localization in chest x-rays using deep
  convolutional neural networks.
\newblock \emph{arXiv preprint arXiv:1705.09850}, 2017.

\bibitem[Kesselman et~al.(2016)Kesselman, Soroosh, Mollura, and
  Group]{kesselman20162015}
Kesselman, Andrew, Soroosh, Garshasb, Mollura, Daniel~J, and Group, RAD-AID
  Conference~Writing.
\newblock 2015 rad-aid conference on international radiology for developing
  countries: The evolving global radiology landscape.
\newblock \emph{Journal of the American College of Radiology}, 13\penalty0
  (9):\penalty0 1139--1144, 2016.

\bibitem[Kingma \& Ba(2014)Kingma and Ba]{Kingma2014}
Kingma, Diederik and Ba, Jimmy.
\newblock Adam: A method for stochastic optimization.
\newblock \emph{arXiv preprint arXiv:1412.6980}, 2014.

\bibitem[Lakhani \& Sundaram(2017)Lakhani and Sundaram]{lakhani2017deep2}
Lakhani, Paras and Sundaram, Baskaran.
\newblock Deep learning at chest radiography: Automated classification of
  pulmonary tuberculosis by using convolutional neural networks.
\newblock \emph{Radiology}, pp.\  162326, 2017.

\bibitem[Mollura et~al.(2010)Mollura, Azene, Starikovsky, Thelwell, Iosifescu,
  Kimble, Polin, Garra, DeStigter, Short, et~al.]{mollura2010white}
Mollura, Daniel~J, Azene, Ezana~M, Starikovsky, Anna, Thelwell, Aduke,
  Iosifescu, Sarah, Kimble, Cary, Polin, Ann, Garra, Brian~S, DeStigter,
  Kristen~K, Short, Brad, et~al.
\newblock White paper report of the rad-aid conference on international
  radiology for developing countries: identifying challenges, opportunities,
  and strategies for imaging services in the developing world.
\newblock \emph{Journal of the American College of Radiology}, 7\penalty0
  (7):\penalty0 495--500, 2010.

\bibitem[Neuman et~al.(2012)Neuman, Lee, Bixby, Diperna, Hellinger, Markowitz,
  Servaes, Monuteaux, and Shah]{neuman2012variability}
Neuman, Mark~I, Lee, Edward~Y, Bixby, Sarah, Diperna, Stephanie, Hellinger,
  Jeffrey, Markowitz, Richard, Servaes, Sabah, Monuteaux, Michael~C, and Shah,
  Samir~S.
\newblock Variability in the interpretation of chest radiographs for the
  diagnosis of pneumonia in children.
\newblock \emph{Journal of hospital medicine}, 7\penalty0 (4):\penalty0
  294--298, 2012.

\bibitem[Potchen et~al.(1979)Potchen, Gard, Lazar, Lahaie, and
  Andary]{Potchen1979}
Potchen, EJ, Gard, JW, Lazar, P, Lahaie, P, and Andary, M.
\newblock Effect of clinical history data on chest film
  interpretation-direction or distraction.
\newblock In \emph{Investigative Radiology}, volume~14, pp.\  404--404, 1979.

\bibitem[Rajpurkar et~al.(2017)Rajpurkar, Hannun, Haghpanahi, Bourn, and
  Ng]{Rajpurkar2017}
Rajpurkar, Pranav, Hannun, Awni~Y, Haghpanahi, Masoumeh, Bourn, Codie, and Ng,
  Andrew~Y.
\newblock Cardiologist-level arrhythmia detection with convolutional neural
  networks.
\newblock \emph{arXiv preprint arXiv:1707.01836}, 2017.

\bibitem[Raoof et~al.(2012)Raoof, Feigin, Sung, Raoof, Irugulpati, and
  Rosenow]{raoof2012interpretation}
Raoof, Suhail, Feigin, David, Sung, Arthur, Raoof, Sabiha, Irugulpati, Lavanya,
  and Rosenow, Edward~C.
\newblock Interpretation of plain chest roentgenogram.
\newblock \emph{CHEST Journal}, 141\penalty0 (2):\penalty0 545--558, 2012.

\bibitem[Wang et~al.(2017)Wang, Peng, Lu, Lu, Bagheri, and Summers]{Wang2017}
Wang, Xiaosong, Peng, Yifan, Lu, Le, Lu, Zhiyong, Bagheri, Mohammadhadi, and
  Summers, Ronald~M.
\newblock Chestx-ray8: Hospital-scale chest x-ray database and benchmarks on
  weakly-supervised classification and localization of common thorax diseases.
\newblock \emph{arXiv preprint arXiv:1705.02315}, 2017.

\bibitem[WHO(2001)]{world2001standardization}
WHO.
\newblock Standardization of interpretation of chest radiographs for the
  diagnosis of pneumonia in children.
\newblock 2001.

\bibitem[Yao et~al.(2017)Yao, Poblenz, Dagunts, Covington, Bernard, and
  Lyman]{Yao2017}
Yao, Li, Poblenz, Eric, Dagunts, Dmitry, Covington, Ben, Bernard, Devon, and
  Lyman, Kevin.
\newblock Learning to diagnose from scratch by exploiting dependencies among
  labels.
\newblock \emph{arXiv preprint arXiv:1710.10501}, 2017.

\bibitem[Zhou et~al.(2016)Zhou, Khosla, Lapedriza, Oliva, and
  Torralba]{Zhou2016}
Zhou, Bolei, Khosla, Aditya, Lapedriza, Agata, Oliva, Aude, and Torralba,
  Antonio.
\newblock Learning deep features for discriminative localization.
\newblock In \emph{Proceedings of the IEEE Conference on Computer Vision and
  Pattern Recognition}, pp.\  2921--2929, 2016.

\end{thebibliography}
\bibliographystyle{icml2017}

\end{document}